\documentclass[10pt,twocolumn,letterpaper]{article}

\usepackage{iccv}
\usepackage{times}
\usepackage{epsfig}
\usepackage{graphicx}
\usepackage{amsmath}
\usepackage{amssymb}

\usepackage[normalem]{ulem}
\usepackage{booktabs}       
\usepackage{array}
\usepackage{multirow}
\usepackage{color}
\usepackage{mathtools}
\usepackage{resizegather}

\usepackage[pagebackref=true,breaklinks=true,letterpaper=true,colorlinks,bookmarks=false]{hyperref}

\iccvfinalcopy 

\def\iccvPaperID{9139} 

\ificcvfinal\fi

\DeclareMathOperator*{\argmax}{argmax}
\DeclareMathOperator*{\argmin}{argmin}

\newcommand{\norm}[1]{\left\lVert#1\right\rVert}
\newcommand\figref{Figure~\ref}
\newcommand\tabref{Table~\ref}

\makeatletter
\def\blfootnote{\gdef\@thefnmark{}\@footnotetext}
\makeatother

\definecolor{forestgreen}{rgb}{0.133, 0.545, 0.133}
\definecolor{brown}{rgb}{0.65, 0.16, 0.16}
\definecolor{pink}{rgb}{0.858, 0.188, 0.478}

\ifx \submission \undefined
  \newcommand{\raquel}[1]{{\color{red}{Raquel: #1}}}
  \newcommand{\james}[1]{\textcolor{blue}{James: #1}}
  \newcommand{\johnson}[1]{\textcolor{forestgreen}{#1}}
  \newcommand{\siva}[1]{\textcolor{pink}{Siva: #1}}
  
  \newcommand{\mengye}[1]{\textcolor{brown}{[MR: #1]}}
  
  \newcommand{\jingkang}[1]{\textcolor{magenta}{[Jingkang: #1]}}
  
\else
  \newcommand{\raquel}[1]{{}}
  \newcommand{\james}[1]{{}}
  \newcommand{\johnson}[1]{{}}
  \newcommand{\siva}[1]{{}}
  
  \newcommand{\mengye}[1]{{}}
  \newcommand{\jingkang}[1]{{}}
  
\fi

\begin{document}


\title{Adversarial Attacks On Multi-Agent Communication}

\author{
  James Tu$^{1,2}$\thanks{Equal contribution.}
  \quad
  Tsunhsuan Wang$^{3*}$
  \quad 
  Jingkang Wang$^{1,2}$
  \quad
  Sivabalan Manivasagam$^{1,2}$
  \\ 
  Mengye Ren$^{1,2}$
  \quad
  Raquel Urtasun$^{1,2}$
  \\
  \\
  $^{1}$Waabi
  \quad
  $^{2}$University of Toronto
  \quad
  $^{3}$MIT
  \\
  \small
  \texttt{\{jtu, wangjk, manivasagam, mren, urtasun\}@cs.toronto.edu}\\
  \small
  \texttt{johnsonwang0810@gmail.com}
  \vspace{-0mm}
}

\maketitle
\ificcvfinal\thispagestyle{empty}\fi

\blfootnote{Work done while all authors were at UberATG.}

\begin{abstract}
Growing at a fast pace, modern autonomous systems will soon be deployed at scale, opening up the possibility for cooperative multi-agent systems. Sharing information and distributing workloads allow autonomous agents to better perform tasks and increase computation efficiency. However, shared information can be modified to execute adversarial attacks on deep learning models that are widely employed in modern systems. 
Thus, we aim to study the robustness of such systems and focus on exploring adversarial attacks in a novel multi-agent setting where communication is done through sharing learned intermediate representations of neural networks.   
We observe that an indistinguishable adversarial message can severely degrade performance, but becomes weaker as the number of benign agents increases. Furthermore, we show that black-box transfer attacks are more difficult in this setting when compared to directly perturbing the inputs, as it is necessary to align the distribution of learned representations with domain adaptation.
Our work studies robustness at the neural network level to contribute an additional layer of fault tolerance to modern security protocols for more secure multi-agent systems. 
\end{abstract}

\vspace{-2mm}
\section{Introduction}
With rapid improvements of modern autonomous systems, it is only a matter of time until they are deployed at scale, opening up the possibility of cooperative multi-agent systems. Individual agents can benefit greatly from shared information to better perform their tasks~\cite{KonecnyMYRSB16,multidrone}.
For example, by aggregating sensory information from multiple viewpoints, a fleet of vehicles can perceive the world more clearly, providing significant safety benefits~\cite{v2vnet}. Moreover, in a network of connected devices, distributed processing across multiple agents can improve computation efficiency~\cite{MobileCloudInference}. While cooperative multi-agent systems are promising, relying on communication between agents can pose security threats as shared information can be malicious or unreliable~\cite{WongS00,borselius2002mobile,NovakRHV03}.

Meanwhile, modern autonomous systems typically rely on deep neural networks known to be vulnerable to adversarial attacks. Such attacks craft small and imperceivable perturbations to drastically change a neural network's behavior and induce false outputs~\cite{szegedy2014intriguing,goodfellow2015explaining,carlini2017towards,madry2017towards}. Even 
if an attacker has
the freedom to send any message, such small perturbations may be the most dangerous as they are indistinguishable from their benign counterparts, making corrupted messages difficult to detect while still highly malicious. 

While modern cyber security algorithms provide adequate protection against communication breaches, adversarial robustness of multi-agent deep learning models has yet to be studied. Meanwhile, when it comes to safety-critical applications like self-driving, additional layers of redundancy and improved security are always welcome. Thus, by studying adversarial robustness, we can enhance modern security protocols by introducing an additional layer of fault tolerance at the neural network level. 

\begin{figure*}[t]
    \centering
    \includegraphics[width=0.95\linewidth]{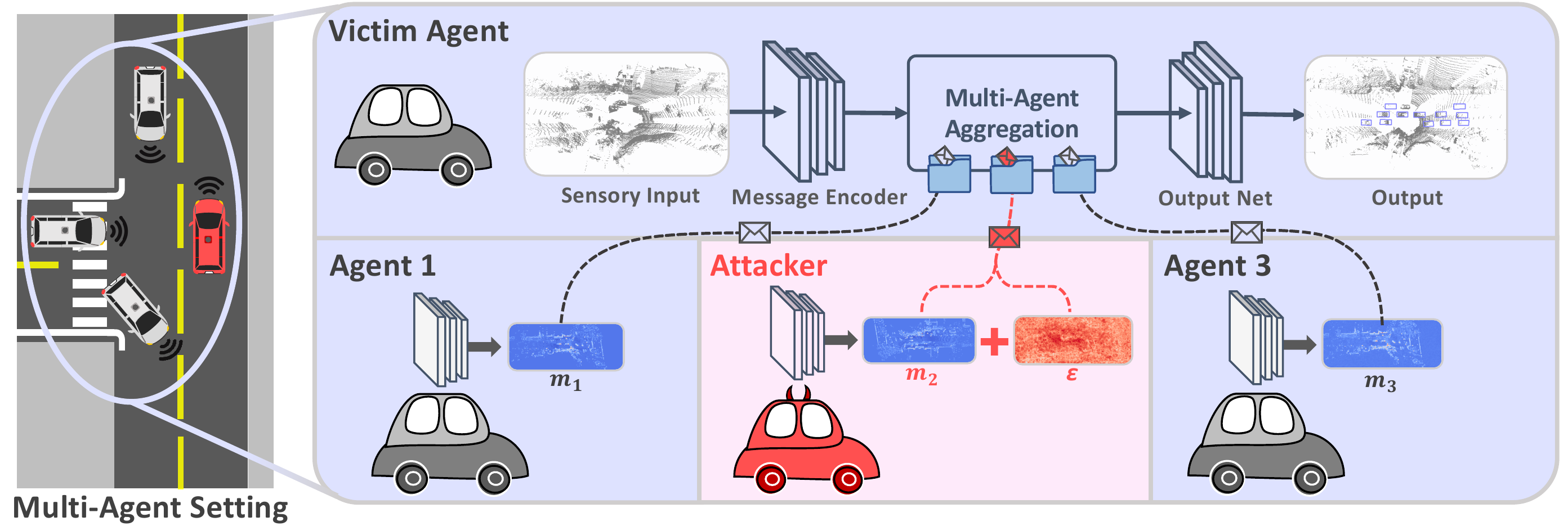}
    \caption{Overview of a multi-agent setting with one malicious agent (red). Here the malicious agent attempts to sabotage a victim agent by sending an adversarial message. The adversarial message is indistinguishable from the original, making the attack difficult to detect.}
    \label{fig:overview}
\end{figure*}

Adversarial attacks have been studied extensively but existing approaches mostly consider attacks on input domains like images~\cite{szegedy2014intriguing,goodfellow2015explaining}, point clouds~\cite{cao2019sensor,advmeshhat}, and text~\cite{SatoSS018,seq2sick}.
On the other hand, multi-agent systems often distribute computation across different devices and transmit intermediate representations instead of input sensory information~\cite{v2vnet,MobileCloudInference}. Specifically, when deep learning inference is distributed across different devices, agents will communicate by transmitting feature maps, which are activations of intermediate neural network layers. Such learned communication has been shown to be superior due to transmitting compact but expressive messages~\cite{v2vnet} as well as efficiently distributing computation~\cite{MobileCloudInference}. 

In this paper, we investigate adversarial attacks in this novel multi-agent setting where perturbations are applied to learned intermediate representations. An illustration is shown in \figref{fig:overview}.
We conduct experiments and showcase vulnerabilities in
two highly practical settings: multi-view perception from images in a fleet of drones and multi-view perception from LiDAR in a fleet of self-driving vehicles (SDVs). 
By leveraging information from multiple viewpoints, these multi-agent systems are able to significantly outperform those that do not exploit communication. 

We show, however, that 
perturbed transmissions which are indistinguishable from the original 
can severely degrade the performance of receivers particularly as the ratio of malicious to benign agents increases.
With only a single attacker, as the number of benign agents increase, attacks become significantly weaker as aggregating more messages decreases the influence of malicious messages. 
When multiple attackers are present, they can coordinate and jointly optimize their perturbations to strengthen the attack.
In terms of defense, when the threat model is known, adversarial training is highly effective, and adversarially trained models can defend against perturbations almost perfectly and even slightly enhance performance on natural examples.
Without knowledge of the threat model, we can still achieve reasonable adversarial robustness by designing more robust message aggregation modules.

We then move on to more practical attacks in a black box setting where the model is  unknown to the adversary.
Since query-based black box attacks need to excessively query a target model that is often unaccessible, we focus on query-free transfer attacks that are more feasible in practice.
However, transfer attacks are much more difficult to execute at the feature-level than on input domains.
In particular, since perturbation domains are model dependent, vanilla transfer attacks
are ineffective because two neural networks with the same functionality can have very different intermediate representations.
Here, we find that training the surrogate model with domain adaptation is key to aligning the distribution of intermediate features and achieve much better transferability.
To further enhance the practicality of attacks, we propose to exploit the temporal consistency of sensory information processed by modern autonomous systems.
When frames of sensory information are collected milliseconds apart, we can exploit the redundancy in adjacent frames to create efficient, low-budget attacks in an online manner.

\section{Related Work}

\begin{figure*}
    \centering
    \includegraphics[width=0.92\linewidth]{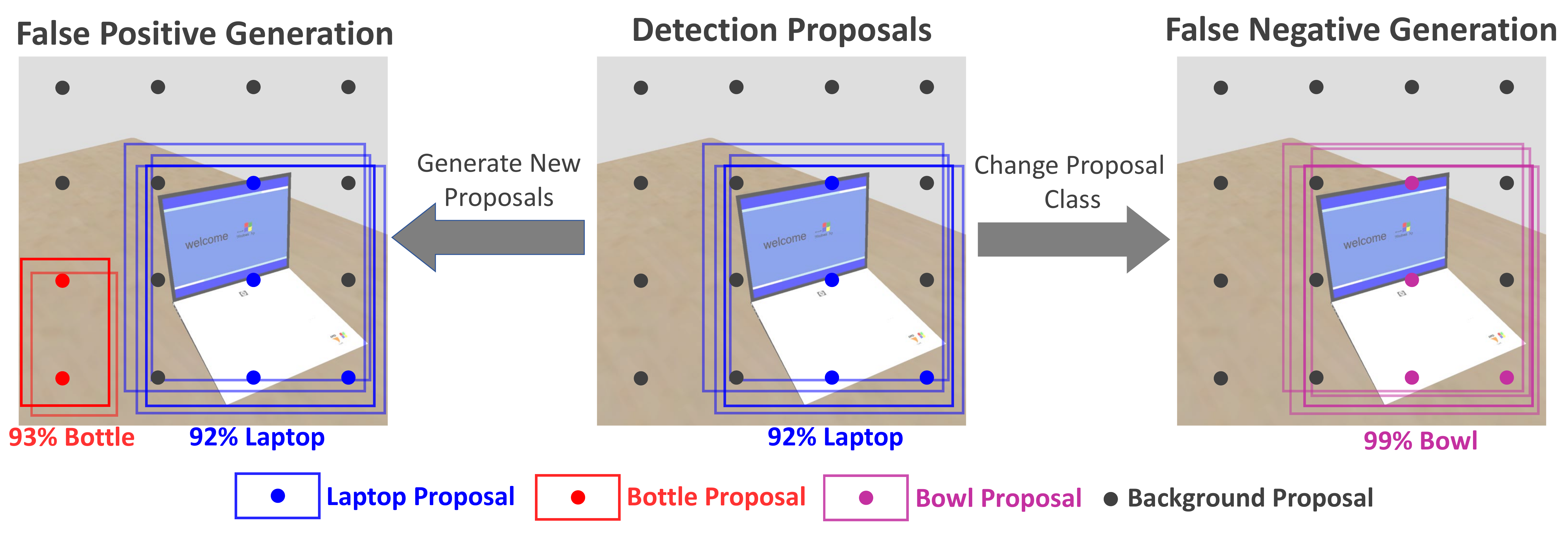}
    \vspace{-2mm}
    \caption{\textbf{Attacking object detection proposals}: False positives are created by changing the class of background proposals and false negatives are created by changing the class of the original proposals.}
    \label{fig:det_attack}
\end{figure*}

\paragraph{Multi-Agent Deep Learning Systems:}
Multi-agent and distributed systems are widely employed in real-world applications to improve computation efficiency~\cite{federated1,DillonWC10,federated2}, collaboration~\cite{v2vnet,multidrone,MobileCloudInference,rauch2012car2x,rockl2008v2v}, and safety~\cite{obst2014multi,nakamoto2019bitcoin}. Recently, autonomous systems have improved greatly with the help of neural networks. New directions have opened up in cooperative multi-agent deep learning systems e.g., federated learning~\cite{federated1,federated2}. 
Although multi-agent communication introduces a multitude of benefits, communication channels are vulnerable to security breaches, as communication channels can be attacked~\cite{comm_attacks_survey, intrusion_det}, encryption algorithms can be broken~\cite{adobe_breach}, and agents can be compromised~\cite{compromised_tesla, compromised_bmw}. 
Thus, imperfect communication channels may be used to execute adversarial attacks which are especially effective against deep learning systems.
While robustness has been studied in the context of federated learning~\cite{ghosh2019robust,BhagojiCMC19,XieHCL20,FangCJG20}, the threat models are different as dataset poisoning and model poisoning are typically used. 
To the best of our knowledge, few works study adversarial robustness on multi-agent deep learning systems during inference. 

\paragraph{Adversarial Attacks:}
Adversarial attacks were first discovered in the context of image classification~\cite{szegedy2014intriguing}, where a small imperceivable perturbation can drastically change a neural network's behaviour and induce false outputs.
Such attacks were then extended to various applications such as semantic segmentation~\cite{xie2017adversarialdetect} and reinforcement learning~\cite{huang2017adversarial}. 
There are two main settings for adversarial attacks - \textit{white box} and \textit{black box}. In a white box setting~\cite{szegedy2014intriguing,goodfellow2015explaining,madry2017towards}, the attacker has full access to the target neural network weights and adversarial examples can be generated using gradient-based optimization to maximize the network's error.
In contrast, black box attacks are conducted without knowledge of the target neural network weights and therefore without any gradient computation. 
In this case, attackers can leverage real world knowledge to inject adversaries that resemble common real world objects~\cite{sun2020towards,nassi2020phantom}. However, if the attacker is able to query the target model, the literature proposes several different strategies to perform query-based attacks ~\cite{boundary_attack,ChenZSYH17,biased_sampling,chen2019hopskipjumpattack}. 
However, query-based attacks are infeasible for some applications as they typically require prohibitively large amounts of queries and computation.
Apart from query-based attacks, a more practical but more challenging alternative is to conduct transfer attacks~\cite{PapernotMGJCS17,xie2019improving,ChengDPSZ19} which do not require querying the target model. 
In this setting, the  attacker trains a surrogate model that imitates the target model. By doing so, the hope is that perturbations generated for the surrogate model will transfer to the target model. 

\vspace{-4mm}

\paragraph{Perturbations In Feature Space:} 
While most works in the literature focus on input domains like images, some prior works have considered perturbations on intermediate representations within neural networks. Specifically,~\cite{JiangMC0J19} estimated the projection of adversarial gradients on a selected subspace to reduce the queries to a target model. \cite{PapernotMSH16,SatoSS018,seq2sick} proposed to generate adversarial perturbation in word embeddings for finding adversarial but semantically-close substitution words. \cite{WuBR17,Zhu2020FreeLB} showed that training on adversarial embeddings could improve the robustness of Transformer-based models for NLP  tasks.

\section{Attacks On Multi-Agent Communication}
This section first introduces the multi-agent framework in which agents leverage information from multiple viewpoints by transmitting intermediate feature maps. 
We then present our method for generating adversarial perturbations in this setting.
Moving on to more practical settings, we consider black box transfer attacks and find that it is necessary to align the distribution of intermediate representations. 
Here, training a surrogate model with domain adaptation can create transferable perturbations.
Finally, we show efficient online attacks by exploiting the temporal consistency of sensory inputs collected at high frequency.

\subsection{Multi-Agent Communication}
We consider a setting where multiple agents cooperate to better perform their tasks by sharing observations from different viewpoints encoded via a learned intermediate representation.
Adopting prior work~\cite{v2vnet}, we assume a homogeneous set of agents using the same neural network. 
Then, each agent $i$ processes sensor input $x_i$ to obtain an intermediate representation $m_i = F(x_i)$.
The intermediate feature map is then broadcasted to other agents in the scene.
Upon receiving messages, agent $j$ will aggregate and process all incoming messages to generate output $Z_j = G(m_1, \dots, m_N)$, where $N$ is the number of agents.
Suppose that an attacker agent $i$ targets a victim agent $j$.
Here, the attacker attempts to send an indistinguishable adversarial message $m_i^{'} = m_i + \delta$ to maximize the error in $Z_j^{'} = G(m_1, \dots m_i + \delta, m_N)$. 
The perturbation $\delta$ is constrained by $\norm{\delta}_p \leq \epsilon$ 
to ensure that the malicious message is subtle and difficult to detect. An overview of the multi-agent setting is shown in \figref{fig:overview}.

\begin{figure*}[t]
    \centering
    \includegraphics[width=.99\linewidth]{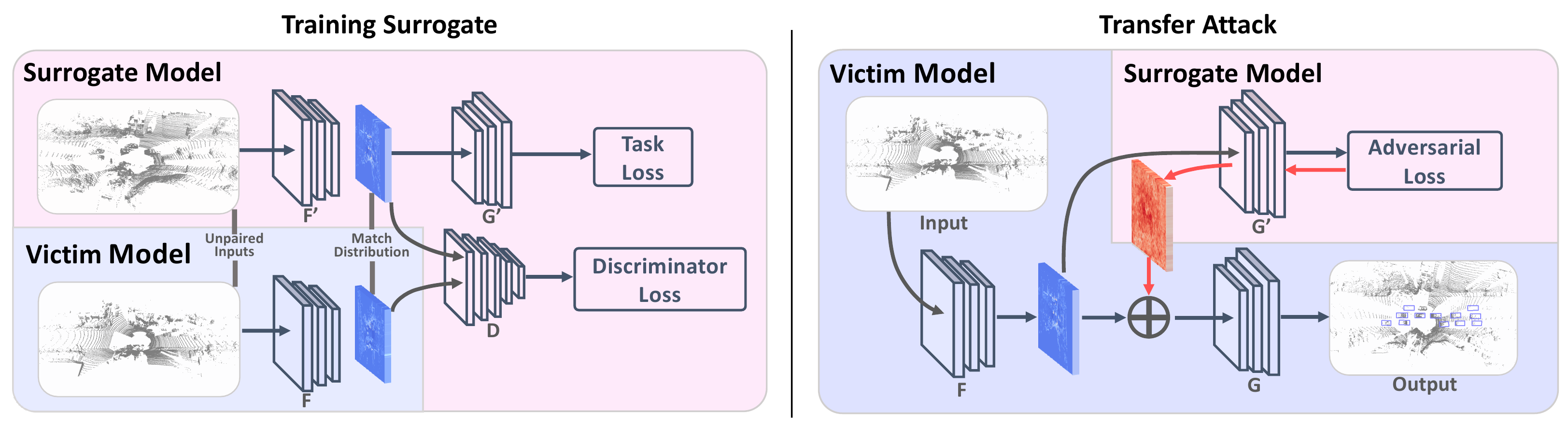}
    \caption{Our proposed transfer attack which incorporates domain adaptation when training the surrogate model. During training, the discriminator forces $F'$ to produce intermediate representations similar to $F$. 
    As a result, $G'$ can generate perturbations that transfer to $G$.
    }
    \label{fig:adda}
\end{figure*}

In this paper, we specifically focus on object detection as it is a challenging task where aggregating information from multiple viewpoints is particularly helpful. In addition, many downstream robotics tasks depend on detection
and thus a strong attack can jeopardize the performance of the full system. 
In this case, output $Z$ is a set of $M$ bounding box proposals ${z^{(1)}, \dots, z^{(M)}}$ at different spatial locations.
Each proposal consists of class scores $z_{\sigma_0}, \dots, z_{\sigma_k}$ and  bounding box parameters describing the spatial location and dimensions of the bounding box.
Here classes $0,\dots,k-1$ are the object classes and $k$ denotes the background class where no objects are detected.

When performing detection, models try to output the correct object class $k$ and maximize the ratio of intersection over union (IOU) of the proposed and ground truth bounding boxes. 
In a post processing step, proposals with high confidence are selected and overlapping bounding boxes are filtered with non-maximum suppression (NMS) to ideally produce a single estimate per ground truth object.

\subsection{Adversarial Perturbation Generation}

We first introduce our loss objective for generating adversarial perturbations against object detection. 
To generate false outputs, we  aim to confuse the proposal class. For detected objects, we suppress the score of the correct class to generate false negatives.
For background classes, false positives are created by pushing up the score of an object class.
In addition, we also aim to minimize the intersection-over-union (IoU) of the bounding box proposals to further degrade performance by producing poorly localized objects.
We define the adversarial loss of the perturbed output $z'$ with respect to an unperturbed output $z$ instead of the ground truth, as it may not always be available to the attacker.
For each proposal $z$, let $u = \argmax_{i} \{z_{\sigma_i} \vert i=0 \dots m\}$ be the highest confidence class. 
Given the original object proposal $z$ and the proposal after perturbation $z'$, our loss function tries to push  $z'$ away from $z$:
\begin{equation}
    \resizebox{1.0\hsize}{!}{$
    \ell_{adv}(z', z) =
    \begin{cases}
    -\log(1 - z'_{\sigma_u}) \cdot \mathrm{IoU}(z', z)  & \text{if } u \neq k \text{ and} \ z_{\sigma_u} > \tau^{+}, \\
    -\lambda \cdot {z'}_{\sigma_v}^{\gamma} \log(1 - z'_{\sigma_v})  & \text{if } u = k \text{ and}\ \ z_{\sigma_u} > \tau^{-}, \\
    0                                       & \text{otherwise}
    \end{cases}
    $}
\end{equation}
An illustration of the attack objective is shown in \figref{fig:det_attack}.
When  $u \neq k$ and the original prediction is not a background class, we apply an untargetted loss to reduce the likelihood of the intended class.
When the intended prediction is the background class $k$, we specifically target a non-background class $v$ to generate a false positive. 
We simply choose  $v$ to be  the class with the  highest confidence that is not the background class.
The $\mathrm{IoU}$ operator denotes the intersection over union of two proposals, $\lambda$ is a weighting coefficient, and 
$\tau^{-}, \tau^{+}$  filter out proposals that are not confident enough. 
We provide more analysis and ablations to justify our loss function design in our experiments. 

Following prior work~\cite{advmeshhat}, it is necessary to minimize the adversarial loss over all proposals. Thus, the optimal perturbation under an $\epsilon$ - $\ell_p$ bound is 
\begin{equation}
    \delta^{\star} =
    \argmin_{\norm{\delta}_p \leq \epsilon}
    \sum_{m=1}^{M} \ell_{adv} (z'^{(m)}, z^{(m)}).
\end{equation}

Our work considers an infinity norm $p=\infty$ and we minimize this loss across all proposals using projected gradient descent (PGD)~\cite{pgd}, clipping $\delta$ to be within $[-\epsilon, \epsilon]$.

\subsection{Transfer Attack}

We also consider transfer attacks as they are the most practical. 
White box attacks assume access to the victim model's weights which is difficult to obtain in practice. On the other hand, query-based optimization is too expensive to execute in real time as state-of-the-art methods still
require thousands of queries 
~\cite{efficientquerybb,HopSkipJumpAttack} on CIFAR-10.  
Instead, when we do not have access to the weights of the victim model $G$, we can imitate it with a surrogate model $G'$ such that perturbations generated by the surrogate model can transfer to the target model. 

\begin{figure*}
    \center
    \includegraphics[width=0.88\linewidth]{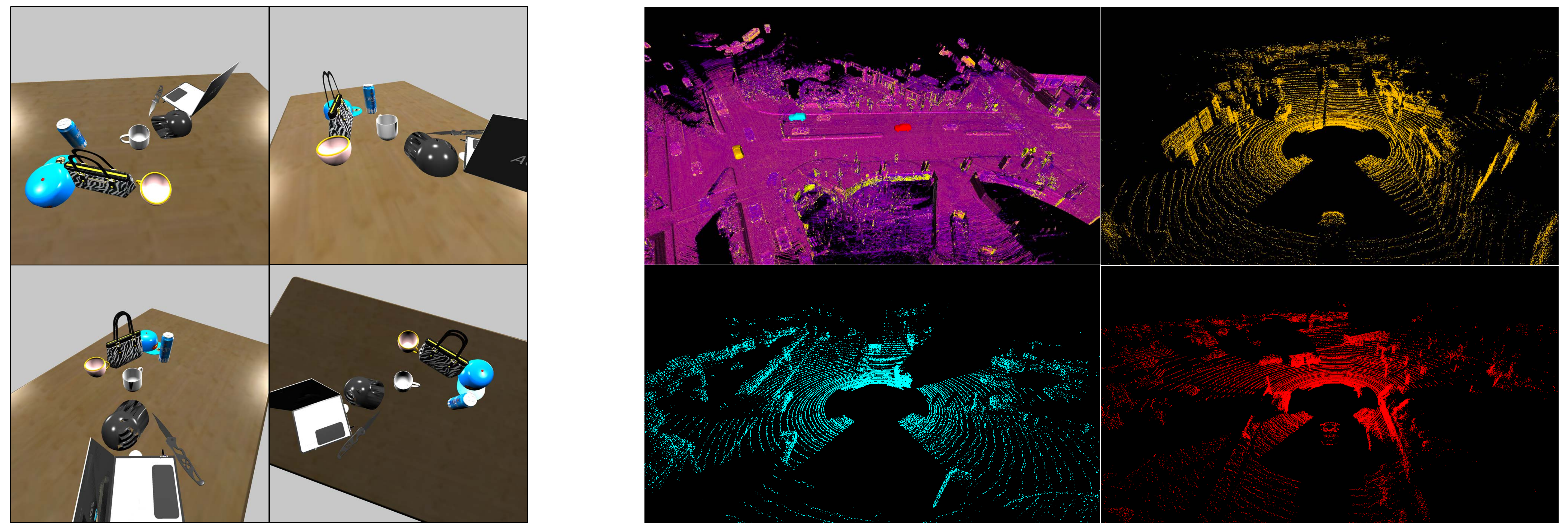}
    \caption{Two multi-agent datasets we use. On the left are images of ShapeNet objects taken from different view points. On the right are LiDAR sweeps by different vehicles in the same scene.}
    \label{fig:datasets}
    \vspace{-2mm}
\end{figure*}

One major challenge for transfer attacks in our setting is that perturbations are generated on intermediate feature maps.
Our experiments show that vanilla transfer attacks are almost completely ineffective
as two networks with the same functionality do not necessarily have the same intermediate representations. 
When training $F$ and $G$, there is no direct supervision on the intermediate features $m=F(x)$. Therefore, even with the same architecture, dataset, 
and training schedule, a surrogate $F'$ may produce messages $m'$ with very different distribution from $m$. 
As an example, a permutation of feature channels carries the same information but results in a different distribution. In general, different random seeds, network initializations or non-deterministic GPU operations
can result in different intermediate representations.
It follows that if $m'$ does not faithfully replicate $m$, we cannot expect $G'$ to imitate $G$.

Thus, to execute transfer attacks, we must have access to samples of the intermediate feature maps.
Specifically, we consider a scenario where the attacker can spy on the victim's communication channel to obtain transmitted messages.  
However, since sensory information is not transmitted, the attacker does not have access to pairs of input $x$ and intermediate representation $m$ to directly supervise the surrogate $F'$ via distillation. 
Thus, we propose to use Adversarial Discriminative Domain Adaptation (ADDA)~\cite{tzeng2017adversarial} to align the distribution of $m$ and $m'$ without explicit input-feature pairs.
An overview is shown in \figref{fig:adda}.

In the original training pipeline, $F'$ and $G'$ would be trained to minimize task loss
\begin{equation}
\small
\mathcal{L}_{task}(z, y, b) =
\begin{cases}
-\log(z_{\sigma_y}) - \mathrm{IoU}(z, b)  & \text{if } y \neq k, \\
-\log(z_{\sigma_y})              & \text{if } y = k, \\
\end{cases}
\end{equation}
where $b$ is a ground truth bounding box and $y$ is its class. The task loss maximizes the log likelihood of the correct class and the IoU between the proposal box and the ground truth box.
In addition, we encourage domain adaptation by introducing a discriminator $D$ to distinguish between real messages $m$ and surrogate messages $m'$.
The three modules $F'$, $G'$, and $D$ can be optimized using the following min-max criterion:
\begin{equation}
    \resizebox{1.02\hsize}{!}{$
    \min\limits_{F'\ G'}\max\limits_{D} 
    \mathcal{L}_{task}(x) +
    \beta \big[ \log D(F(x)) +  \log (1 - D(F'(x)))]
    $}
\end{equation}
where $\beta$ is a weighting coefficient and 
we use binary cross entropy loss to supervise the discriminator.
During training, we adopt spectral normalization~\cite{miyato2018spectral} in the discriminator and the two-time update rule~\cite{heusel2017gans} for stability.

\subsection{Online Attack}
\label{sec:online_attack}
In modern applications of autonomous systems, consecutive frames of sensory information are typically collected only milliseconds apart. 
Thus, there is a large amount of redundancy between consecutive frames which can be exploited to achieve more efficient adversarial attacks.
Following previous work~\cite{wei19video} in images, we propose to exploit this redundancy by using the perturbation from the previous time step as initialization for the current time step.

Furthermore, we note that intermediate feature maps capture the spatial context of sensory observations, which change due to the agent's egomotion.
Therefore, by applying a rigid transformation on the perturbation at every  time step to account for egomotion, we can generate stronger perturbations that are synchronized with the movement of sensory observations relative to the agent.
In this case, the perturbations are  updated as follows:
\begin{equation}
\delta^{(t+1)} \leftarrow H_{t \rightarrow t+1} (\delta^{(t)}) \ -  \alpha \nabla_{H_{t \rightarrow t+1}(\delta)} \mathcal{L}_{adv}(Z'^{(t+1)}, Z^{(t+1)}).
\end{equation}
Here $H_{t \rightarrow t+1}$ is a rigid transformation mapping the attacker's pose at time $t$ to $t+1$ and $\alpha$ is the step size. 
By leveraging temporal consistency we can generate strong perturbations with only one gradient update per time step, making online attacks more feasible.

\section{Experiments}
\label{sec:experiments}
\subsection{Multi-Agent Settings}

\begin{figure}
    \centering
        \includegraphics[width=0.95\linewidth]{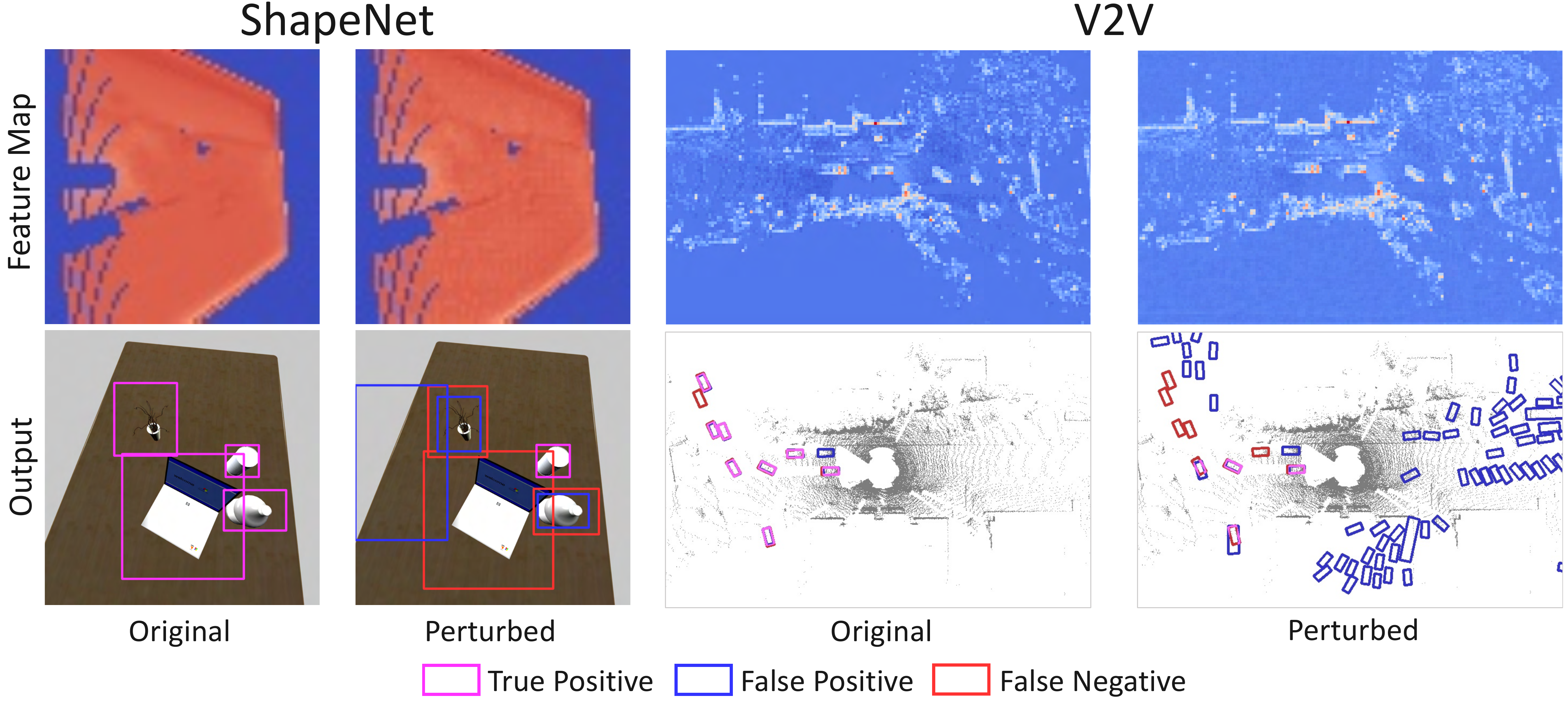}
    \caption{Qualitative attack examples. Top: Messages sent by another agent visualized in bird's eye view. Bottom: outputs. Perturbations are very subtle but severely degrade performance. 
    }
   \label{fig:qua}
\end{figure}

\paragraph{Multi-View ShapeNet:}
We conduct our attacks on multi-view detection from images, which is a common task for a fleets of drones. Following prior work~\cite{ricson}, we generate a synthetic dataset by placing 10 classes of ShapeNet~\cite{shapenet} objects on a table (see \figref{fig:datasets}). 
From each class, we subsample 50 meshes and use a 40/10 split for training and validation. In every scene, we place 4 to 8 objects and perform collision checking to ensure objects do not overlap. Then, we capture 128$\times$128 RGB-D images from  2 to 7 viewpoints sampled from the upper half of a sphere centered at the table center with a radius of 2.0 units.
This dataset consists of 50,000 training scenes and 10,000 validation scenes.
When conducting attacks, we randomly sample one of the agents to be the adversary.
Our detection model uses an architecture similar to the one introduced in \cite{ricson}.
Specifically, we process input RGB-D images using a U-Net~\cite{unet} and then unproject the features into 3D using the depth measures. 
Features from all agents are then warped into the same coordinate frame and aggregated with mean pooling.
Finally, aggregated features are processed by a 3D U-Net and a detection header to generate 3D bounding box proposals.

\paragraph{Vehicle To Vehicle Communication:}
We also consider a self-driving setting with vehicle-to-vehicle(V2V) communication. Here, we adopt the dataset used in ~\cite{v2vnet}, where 3D reconstructions of logs of real world LiDAR scans are simulated from the perspectives of other vehicles in the scene  using a high-fidelity LiDAR simulator~\cite{lidarsim}. 
These logs are collected by self-driving vehicles equipped with LiDAR sensors capturing 10 frames per second (see \figref{fig:datasets}).
The training set consists of 46,796 subsampled frames from the logs and we do not subsample the validation set, resulting in 96,862 frames.
In every log we select one attacker vehicle and sample others to be cooperative agents with up to 7 agents in each  frame unless otherwise specified. 
This results in a consistent assignment of attackers and V2V agents throughout the frames.
In this setting, we use the state-of-the-art perception and motion forecasting model V2VNet ~\cite{v2vnet}.
Here, LiDAR inputs are first encoded into bird's eye view (BEV) feature maps.
Feature maps from all agents are then warped into the ego coordinate frame and aggregated with a GNN to produce BEV bounding box proposals. More details of the ShapeNet model and V2VNet are provided in the supplementary material.

\begin{figure}
    \centering
    \includegraphics[width=1.0\linewidth]{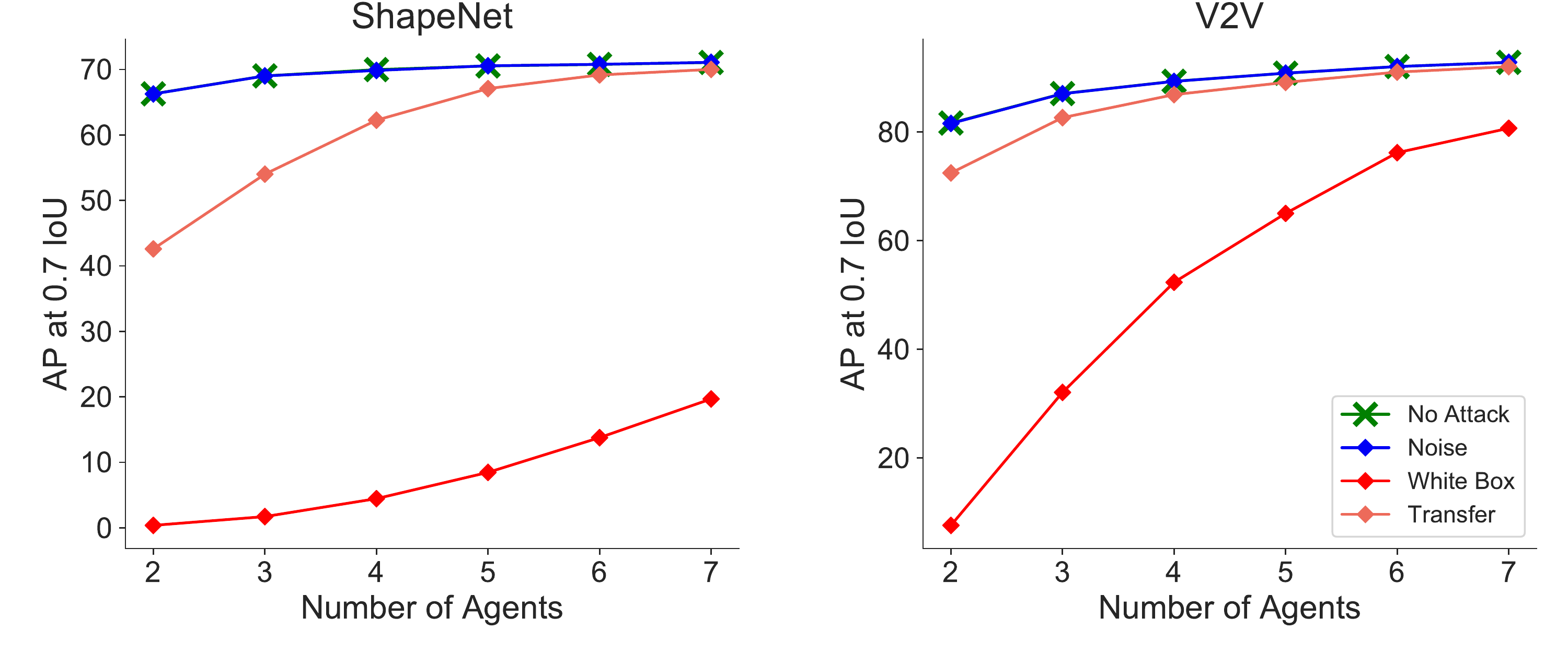}
    \caption{Evaluation under no perturbation, uniform noise, transfer attack, and white box attack. Results are grouped by the number of agents in the scene where one agent is the attacker.}
    \label{fig:main_attacks}
\end{figure}

\paragraph{Implementation Details:}
When conducting attacks, we set $\epsilon=0.1$.
For the proposed loss function, we set $\lambda = 0.2, \tau^{-} = 0.7, \tau^{+}=0.3$, and $\gamma = 1$. 
Projected gradient descent is done using Adam with learning rate $0.1$ 
and we apply $15$ PGD steps for ShapeNet and only $1$ PGD step for low budget online attacks in the V2V setting.  
The surrogate models use the same architecture and dataset as the victim models. When training the surrogate model, we set $\beta=0.01$, model learning rate $0.001$, and discriminator learning rate $0.0005$.
For evaluation, we compute area under the precision-recall curve of bounding boxes, where bounding boxes are correct if they have an IoU greater than 0.7 with a ground truth box of the same class. We refer to this metric as \textit{AP at 0.7} in the following.  

\subsection{Results}

\paragraph{Attack Results:}
Visualizations of our attack are shown in \figref{fig:qua} and we present quantitative results of our attack and baselines in \figref{fig:main_attacks}.
We split up the evaluation by the number of agents in the scene and one of the agents is always an attacker. 
As a baseline, we sample the perturbation from $\mathcal{U}{(-\epsilon, \epsilon)}$ to demonstrate that the same $\epsilon$ bounded uniform noise does not have any impact on detection performance.
The white box attack is especially strong when few agents are in the scene, but becomes weaker as the number of benign agents increase, causing the relative weight of the adversarial features in mean pooling layers to decrease.
Finally, our transfer attack with domain adaptation achieves moderate success with few agents in the scene, but is significantly weaker than the white box attack.

\begin{table}
    \centering
    \resizebox{0.99\linewidth}{!}{
    \begin{tabular}{@{} lcccc  @{}}
        \toprule 
        \multirow{3}{2.5cm}{\centering } & \multicolumn{2}{c}{ShapeNet} & \multicolumn{2}{c}{V2V} \\
        \cmidrule{2-3} \cmidrule{4-5}
                                  & Clean & Perturbed & Clean & Perturbed \\
        \midrule                         
        Original             & 66.33 & 0.62 & 82.19 & 7.55 \\
        Adv Trained          & \textbf{67.29} & \textbf{66.00} & \textbf{82.60} & \textbf{83.44} \\
        \bottomrule
    \end{tabular}
    }
    \vspace{2mm}
    \caption{Results of adversarial training. Robustness increases significantly, matching clean inference. Furthermore performance on clean data also improves slightly.
    }
    \label{tab:adv_train}
\end{table}

\paragraph{Robustifying Models:}
To defend against our proposed attack, we conduct adversarial training against the white box adversary and show the results in \tabref{tab:adv_train}. Here, we follow the standard adversarial training set up, except perturbations are applied to intermediate features instead of inputs. This objective can be formulated as 
\begin{multline}
\min_{\theta} \mathbb{E}_{(x,y) \sim D}  
    \max_{\left\lVert \delta \right\rVert_{\infty} < \epsilon} \phi((x,y, \delta); \theta) := \\ \mathcal{L}_{task}\left(G(F(x_0), \dots, F(x_i) + \delta, \dots, F(x_N); \theta)\right),
\end{multline}
where  $D$ is the natural training distribution and $\theta$ denotes model parameters. During training, we generate a new perturbation $\delta$ for each training sample. 
In the multi-agent setting, we find it easier to recover from adversarial perturbations when compared to traditional single-agent attacks. 
Moreover, adversarial training is able to slightly improve performance on clean data as well, while adversarial training has been known to hurt natural performance in previous settings~\cite{advmixup,tsipras2018robustness}.

\begin{table}
    \centering
    \resizebox{1.0\linewidth}{!}{
    \begin{tabular}{@{} lcccccc  @{}}
        \toprule 
         & \multicolumn{3}{c}{Clean} & \multicolumn{3}{c}{Perturbed} \\
        \cmidrule{2-4} \cmidrule{5-7}
        Agents  & 2  & 4 & 6  & 2 & 4 & 6 \\
        \midrule                         
        Mean Pool       & 82.09 & 89.25 & 92.43 & 0.90 & 12.93 & 41.77 \\
        GNN(Mean)       & \textbf{82.19} & \textbf{89.93} & \textbf{92.94} & 7.55 & 52.31 & 76.18 \\
        GNN(Median)     & 82.11 & 87.12 & 90.75 & 12.8 & \textbf{67.70} & \textbf{86.30} \\
        GNN(Soft Med)   & 82.19 & 89.67 & 92.49 & \textbf{21.53} & 61.37 & 84.99 \\
        \bottomrule
    \end{tabular}
    }
    \vspace{2mm}
    \caption{Choice of fusion in V2VNet affects performance and robustness. We investigate using mean pooling and using a GNN with various aggregation methods.}
    \label{tab:which_aggr}
\end{table}

While adversarial training is effective in this setting, it requires knowledge of the threat model. When the threat model is unknown, we can still naturally boost the robustness of multi-agent models with the design of the aggregation module. Specifically, we consider several alternatives to V2VNet's GNN fusion and present the performance under attacked and clean data in \tabref{tab:which_aggr}. First, replacing the entire GNN with an adaptive mean pooling layer significantly decreases robustness. On the other hand, we swap out the mean pooling in GNN nodes with median pooling and find that it increases robustness at the cost of performance on clean data with more agents, since more information is discarded. We refer readers to the supplementary materials for more details on implementation of the soft median pooling.

\paragraph{Multiple Attackers:}
We previously focused on settings with one attacker, and now conduct experiments with multiple attackers in the V2V setting.
In each case, we also consider if attackers are able to cooperate. In cooperation, attackers jointly optimize their perturbations. Without cooperation, attackers are blind to each other and optimize their perturbations assuming other messages have not been perturbed.  
Results with up to 3 attackers are shown in \tabref{tab:multi_atk}. 
As expected, more attackers can increase the strength of attack significantly, furthermore, if multiple agents can coordinate, a stronger attack can be generated.

\begin{table}
    \centering
    \resizebox{1.03\linewidth}{!}{
    \begin{tabular}{@{} lcccccc  @{}}
        \toprule 
         & \multicolumn{3}{c}{Cooperative} & \multicolumn{3}{c}{Non-Cooperative} \\
        \cmidrule{2-4} \cmidrule{5-7}
        Agents  & 4  & 5 & 6  & 4 & 5 & 6 \\
        \midrule                         
        1 Attacker   & 52.31 & 65.00 & 76.18 & 52.31 & 65.00 & 76.18 \\
        2 Attacker   & 28.31 & 41.34 & 54.50 & 39.02 & 51.96 & 64.02 \\
        3 Attacker   & \textbf{12.07} & \textbf{22.84} & \textbf{35.13} & \textbf{24.27} & \textbf{38.17} & \textbf{51.58} \\
        \bottomrule
    \end{tabular}
    }
    \vspace{2mm}
    \caption{Multiple white box attackers in the V2V setting. Cooperative attackers jointly optimize their perturbations and non-cooperative attackers optimize without knowledge of each other.}
    \label{tab:multi_atk}
\end{table}

Next, we apply adversarial training to the multi-attacker setting and present results in \tabref{tab:multi_adv_tr}. Here, all attacks are done in the cooperative setting and we show results with 4 total agents. Similar to the single attacker setting, adversarial training is highly effective. However, while adversarial training against one attacker improves performance in natural examples, being robust to stronger attacks sacrifices performance on natural examples. This suggests that adversarial training has the potential to improve general performance when an appropriate threat  model is selected. Furthermore, we can see that training on fewer attacks does not generalize perfectly to more attackers but the opposite is true. Thus, it is necessary to train against an equal or greater threat model to fully defend against such attacks.

\begin{table}
    \centering
    \resizebox{0.75\linewidth}{!}{
    \begin{tabular}{@{} lcccc  @{}}
        \toprule 
        Attackers  & 0 & 1 & 2 & 3  \\
        \midrule                         
        Train On 0 & 89.93 & 52.31 & 28.31 & 12.07 \\
        Train On 1 & \textbf{90.09} & \textbf{90.00} & 81.95 & 75.28 \\
        Train On 2 & 89.71 & 89.68 & 88.91 & 88.33 \\
        Train On 3 & 89.55 & 89.51 & \textbf{88.94} & \textbf{88.51} \\
        \bottomrule
    \end{tabular}
    }
    \vspace{2mm}
    \caption{Adversarial training with multiple attackers in the V2V setting. We train on settings with various number of attackers and evaluate the models across the settings.} 
    \label{tab:multi_adv_tr}
\end{table} 

\begin{figure}[t]
    \centering
    \resizebox{1.0\linewidth}{!}{
    \begin{tabular}{>{\centering}m{1.05cm} m{\linewidth} @{}}
        Victim & \includegraphics[width=1.0\linewidth]{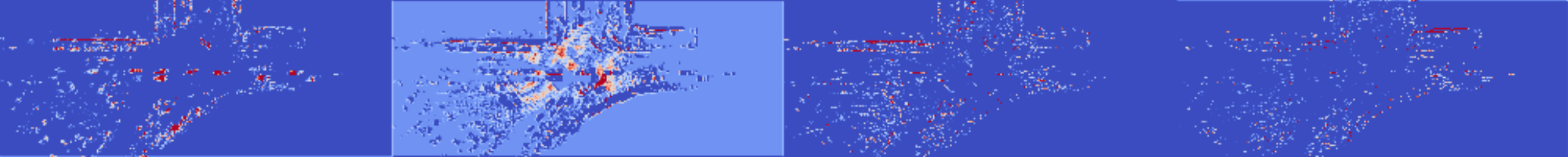} \\
        DA & \includegraphics[width=1.0\linewidth]{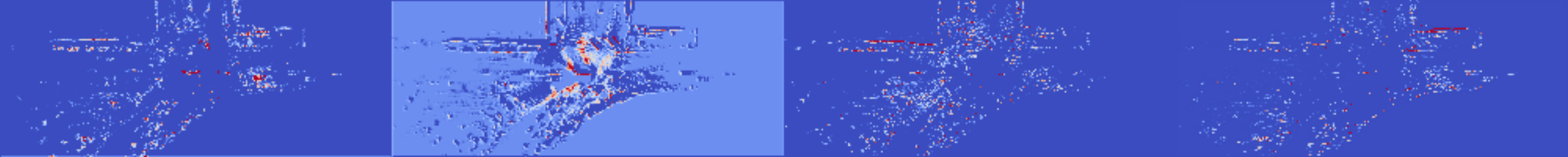} \\
        No DA &\includegraphics[width=1.0\linewidth]{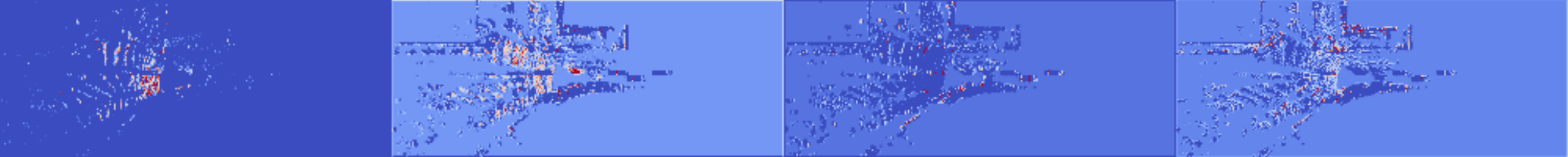} \\
    \end{tabular}
    }
    \vspace{2mm}
\caption{Visualization of how domain adaptation(DA) affects 4 channels of the intermediate feature map. Observe that the surrogate trained with DA closely imitates the victim model, while the surrogate trained without DA produces different features.}
\label{fig:feature_vis}
\end{figure}

\paragraph{Domain Adaptation:}
More results of the transfer attack are included in \tabref{tab:xfer}. First, we conduct an ablation and show that a transfer attack without domain adaptation (DA) is almost completely ineffective.
On the contrary,  surrogate models trained with DA achieve significant improvements.
A visual demonstration of feature map alignment with DA is shown in \figref{fig:feature_vis}, visualizing 4 channels of the intermediate feature maps. Features from a surrogate trained with DA is visually very similar to the victim, while a surrogate trained without DA produces features with no resemblance.

Since our proposed DA improves the transferability of the surrogate model, we can further improve our transfer attack by also adopting methods from the literature which enhance the transferability of a given perturbation. We find that generating perturbations from diversified inputs (DI)~\cite{xie2019improving} is ineffective as resizing input feature maps distorts spatial information which is important for localizing objects detection. On the other hand, using an intermediate level attack projection (ILAP)~\cite{Huang2019} yields a small improvement. Overall, we find transfer attacks more challenging when at the feature level. In standard attacks on sensory inputs, perturbations are transferred into the same input domain. However, at a feature level the input domains are model-dependent, making transfer attacks between different models more difficult. 
\begin{table}
    \centering
    \resizebox{0.85\linewidth}{!}{
    \begin{tabular}{@{} l@{\hskip 1cm}c@{\hskip 1cm}c  @{}}
        \toprule 
                                  & ShapeNet   & V2V \\
        \midrule                         
        Clean                     & 66.28 & 82.19 \\
        Transfer                  & 66.21 & 81.31  \\
        Transfer + DA             & 42.59 & 72.45  \\
        Transfer + DA + ILAP      & \textbf{35.69} & \textbf{71.76}  \\
        Transfer + DA + DI        & 49.38 & 75.18  \\
        \bottomrule
    \end{tabular}
    }
    \vspace{2mm}
    \caption{Transfer attacks evaluated with 2 agents. Training the surrogate with domain adaptation (DA) significantly improves transferability. In addition, we attempt to enhance transferability with ILAP~\cite{Huang2019} and DI~\cite{xie2019improving}.}
    \label{tab:xfer}
\end{table}

\vspace{-0.1in}
\paragraph{Online Attacks:}
We conduct an ablation on the proposed methods for exploiting temporal redundancy in an online V2V setting, shown in \tabref{tab:online_ablation}. 
First, if we ignore temporal redundancy and do not reuse the previous perturbation, attacks are much weaker. In this evaluation we switch from PGD~\cite{pgd} to FGSM~\cite{goodfellow2015explaining} 
to obtain a stronger perturbation in one update for fair comparison.
We also show that applying a rigid transformation on the perturbations at every frame to compensate for egomotion provides a modest improvement to the attack when compared to the \textit{No Warp} ablation. 

\vspace{-0.1in}
\paragraph{Loss Function Design:}
We conduct an ablation study on using our adversarial loss $\mathcal{L}_{adv}$ instead of the negative task loss $-\mathcal{L}_{task}$
in \tabref{tab:loss_ablation}. 
This ablation validates our loss function and showcase that for structured outputs, properly designed adversarial losses is more effective than the naive negative task loss which is widely used in image classification tasks. 
Our choice for the loss function design is motivated by our knowledge of the post-processing non-maximum suppression (NMS).
Since NMS selects bounding boxes with the highest confidence in a local region, proposals with higher scores should receive stronger gradients. 
More specifically, an appropriate loss function of $f$ for proposal score $\sigma$ should satisfy
$
   \left( \vert \nabla_{\sigma_2} f(\sigma_2)\vert - \vert \nabla_{\sigma_1} f(\sigma_1)\vert \right) /
    {(\sigma_2 - \sigma_1)} > 0
$
so that $\vert \nabla_{\sigma} f(\sigma) \vert$ is monotonically increasing in $\sigma$. We can see that the standard log likelihood does not satisfy this criteria, which explains why our loss formulation is more effective.
In addition, we add the focal loss term~\cite{focalloss}
to generate more false positives, as aggressively focusing on one proposal in a local region is more effective due to NMS.

\begin{table}
    \centering
    \setlength\tabcolsep{4.5pt}
    \resizebox{0.8\linewidth}{!}{
    \begin{tabular}{@{} l  c  c  c @{}}
    \toprule
    AP @ 0.7 & 2 Agents & 4 Agents & 6 Agents \\
    \midrule
    Our Attack & 7.55 & \textbf{52.31} & \textbf{76.18} \\
    No Warping & \textbf{7.17} & 52.35 & 77.37 \\
    Independent & 56.98 & 80.21 & 87.05 \\
    \bottomrule
    \end{tabular}
    }
    \vspace{2mm}
    \caption{
    Ablation on online attacks in the V2V setting. \textit{Independent} refers to treating each frame independently and not reusing previous perturbations. \textit{No warp} refers to omitting the rigid transformation to account for egomotion.
    }
    \label{tab:online_ablation}
\end{table}

\begin{table}[t]
    \centering
    \setlength\tabcolsep{4.5pt}
    \resizebox{0.90\linewidth}{!}{
    \begin{tabular}{@{} l  l  c  c  c  @{}}
    \toprule
    \multicolumn{2}{c}{} & 2 Agents & 4 Agents & 6 Agents \\
    \midrule
    \multirow{2}{*}{ShapeNet} & $-\mathcal{L}_{task}$ & 6.10 & 20.07 & 29.00 \\
                              & $\mathcal{L}_{adv}$ & \textbf{0.37} & \textbf{4.45} & \textbf{13.77} \\
    \midrule
    \multirow{2}{*}{V2V} & $-\mathcal{L}_{task}$ & 20.8 & 63.82 & 79.11 \\
                         & $\mathcal{L}_{adv}$ & \textbf{7.55} & \textbf{52.31} & \textbf{76.18} \\
    \bottomrule 
    \end{tabular}
    }
    \vspace{2mm}
    \caption{Ablation on loss function, it produces stronger adversarial attacks than simply using the negative of the training task loss. 
    }
    \label{tab:loss_ablation}
\end{table}

\section{Conclusion}
In this paper, we investigate adversarial attacks on communication in multi-agent deep learning systems. 
Our experiments in two practical settings demonstrate that compromised communication channels can be used to execute adversarial attacks. However, robustness increases as the ratio of benign to malicious actors increases. 
Furthermore, we found that more practical transfer attacks are more challenging in this setting and require aligning the distributions of intermediate representations.
Finally, we propose a method to achieve efficient and practical online attacks by exploiting temporal consistency of sensory inputs.
We believe studying adversarial robustness on multi-agent deep learning models in real-world applications is an important step towards more secure multi-agent systems.

{\small
\bibliographystyle{ieee_fullname}
\bibliography{bib}
}

\end{document}